\documentclass[runningheads]{llncs}
\usepackage{eccv}

\usepackage{eccvabbrv}

\usepackage{graphicx}
\usepackage{booktabs}

\usepackage[accsupp]{axessibility}  % Improves PDF readability for those with disabilities.

\usepackage{hyperref}

\usepackage{orcidlink}

\begin{document}

\title{MM-Diff: High-Fidelity Image Personalization via Multi-Modal Condition Integration} 

\titlerunning{MM-Diff: Image Personalization via Multi-Modal Condition Integration}

\author{Zhichao Wei \and Qingkun Su \and Long Qin \and Weizhi Wang}

\authorrunning{Z.~Wei et al.}

\institute{Alibaba Group}

\maketitle

\begin{abstract}
  Recent advances in tuning-free personalized image generation based on diffusion models are impressive. However, to improve subject fidelity, existing methods either retrain the diffusion model or infuse it with dense visual embeddings, both of which suffer from poor generalization and efficiency. Also, these methods falter in multi-subject image generation due to the unconstrained cross-attention mechanism. In this paper, we propose MM-Diff, a unified and tuning-free image personalization framework capable of generating high-fidelity images of both single and multiple subjects in seconds. Specifically, to simultaneously enhance text consistency and subject fidelity, MM-Diff employs a vision encoder to transform the input image into CLS and patch embeddings. CLS embeddings are used on the one hand to augment the text embeddings, and on the other hand together with patch embeddings to derive a small number of detail-rich subject embeddings, both of which are efficiently integrated into the diffusion model through the well-designed multimodal cross-attention mechanism. Additionally, MM-Diff introduces cross-attention map constraints during the training phase, ensuring flexible multi-subject image sampling during inference without any predefined inputs (\eg, layout). Extensive experiments demonstrate the superior performance of MM-Diff over other leading methods.
\end{abstract}

\section{Introduction}
\label{sec:intro}

\begin{figure}[tb]
  \centering
  \includegraphics[width=12.0cm]{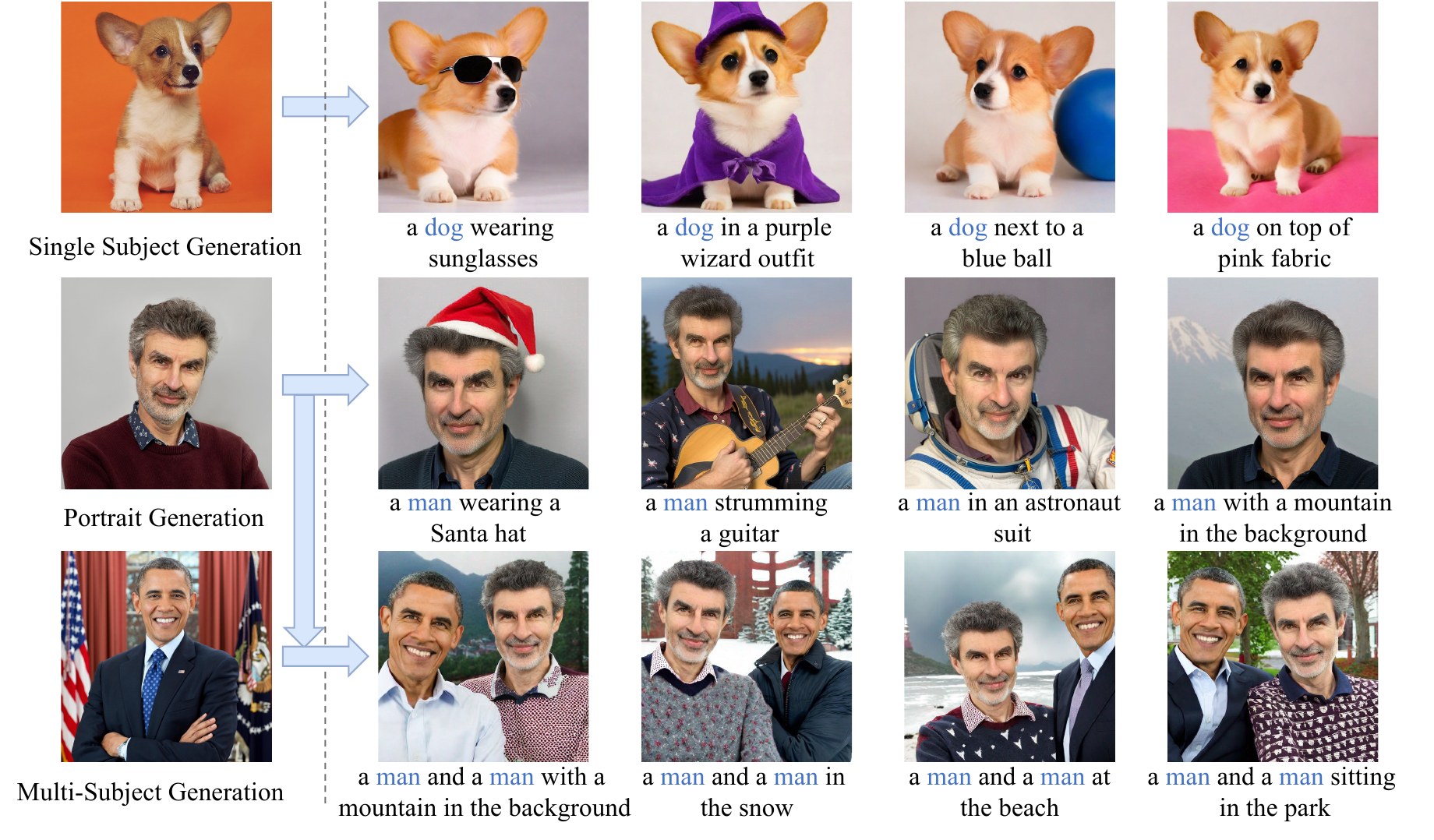}
  \caption{Given a single reference image, our MM-Diff can generate diverse personalized images guided by the text prompt in seconds. Moreover, our model supports multi-subject image generation without any predefined inputs (\eg, layouts).
  }
  \label{fig:intro}
\end{figure}

Personalized image generation \cite{ruiz2023dreambooth, gal2022image, kumari2023multi, wei2023elite} aims to render a subject in novel scenes, styles and actions. In contrast to text-to-image generation \cite{nichol2021glide, ramesh2022hierarchical, rombach2022high, pernias2023wuerstchen} that generate random images based on the provided text prompts, personalized image generation offers users the capability to customize outputs with an additional focus on the given subject, giving rise to wider applications, such as picture book generation \cite{wang2023autostory} and portrait generation \cite{li2023photomaker}.

With the evolution of diffusion-based text-to-image methods \cite{rombach2022high, betker2023improving, pernias2023wuerstchen} recently, many diffusion-based personalized image generation methods have emerged to meet users' customized demand. The most commonly used method in the field is fine-tuning-based methods \cite{ruiz2023dreambooth, kumari2023multi, han2023svdiff, hao2023vico}, which require several images of the specified subject to fine-tune the model. While the images generated by these methods demonstrate high consistency with the provided subject, they suffer from two key limitations: cost of per-subject personalization and inflexibility in generating images of multiple subjects. Personalization is expensive, as these methods typically need 10-30 minutes to fine-tune the model for each new subject using specially crafted data, consuming substantial computational resources and human effort. Moreover, existing methods are inflexible in generating images with multiple subjects, often requiring manually designed layouts or rules to tackle the attribute binding issue \cite{gu2024mix}, where the model confuses the characteristics of different subjects. To reduce the difficulty of data collection and accelerate subject personalization, recent tuning-free works \cite{xiao2023fastcomposer, wei2023elite, ma2023subject, ye2023ip, li2023photomaker}, trained on large-scale domain-specific or open-domain datasets, have employ a visual encoder to embed the input reference image. Once trained, these models can encode any image from the pretrained domain into embeddings and achieve personalized generation through few steps of fine-tuning or even without any tuning process. However, existing methods enhance subject fidelity by either training the entire weights of the base model or by incorporating dense visual features extracted by the encoder into the base model. The former approach suffers from issues of poor generalization, while the latter imposes high computational burden. Meanwhile, these methods still struggle with multi-subject generation due to insufficient constraints on the model optimization process.

Building on the analysis presented, we propose the MM-Diff, a unified personalized image generation framework that enables rapid single- and multi- subject personalization without any fine-tuning process. To enhance subject fidelity, our key idea is to inject a small number of subject embeddings containing rich subject details into the diffusion process. We use a vision encoder to derive the preliminary forms of these subject embeddings from the reference image and introduce a Subject Embedding Refiner (SE-Refiner) to enrich the details of subject embeddings using the patch-level visual embeddings extracted by the same encoder. We also fuse the generic entity token in text embeddings, such as "man", with the global subject representations derived by the vision encoder to further improving text consistency and subject fidelity. The resultant visual embeddings, coupled with the augmented text embeddings, are seamlessly integrated into the diffusion model through the innovative deployment of LoRA layers \cite{hu2021lora} on the key and value projections in cross-attention mechanism. Our design is generic across any data domain and allows the rapid generation of high-fidelity images without any fine-tuning.

For multi-subject image generation, previous works \cite{xiao2023fastcomposer, ma2023subject, gu2024mix} have identified the attribute binding issue as a consequence of the unrestricted cross-attention mechanism, wherein each entity token in text embeddings could attend not only to the image region of its corresponding subject but also to the image regions of other subjects, making the model confused to distinguish between different subjects. Following \cite{xiao2023fastcomposer}, we introduce cross-attention map constraints to tackle this problem, while also taking into account the injection of subject embeddings. The essence of the proposed constraints is to guide the model to associate different entity tokens in both text and image condition with distinct, non-overlapping regions of the image during training phrase. This strategy facilitates the generation of high-quality multi-subject images during inference phase without the need for additional inputs, such as a predefined layout.

As demonstrated in \cref{fig:intro}, the proposed MM-Diff is capable of accomplishing both single- and multi- subject personalization across various domains in seconds, delivering impressive subject fidelity and text consistency. In terms of quantitative comparisons, our method either surpasses or is on par with state-of-the-art (SOTA) methods, including fine-tuning and tuning-free methods, on various testsets. To summarize, our contributions are as follows:
\begin{itemize}
    \item[$\bullet$] We present a tuning-free personalized image generation framework, dubbed MM-Diff, for fast and high-fidelity images generation. It integrates vision-augmented text embeddings and a small number of detail-rich subject embeddings into the diffusion model via the proposed multi-modal cross-attention mechanism, improving text and subject consistency efficiently.
    \item[$\bullet$] We introduce the cross-attention map constraints to tackle the attribute binding issue in multi-subject personalization. These constraints guide the model to associate entity tokens in both text and image condition with distinct image regions during training, thus allowing for flexible multi-subject image generation during inference without any predefined inputs.
    \item[$\bullet$] Both quantitative and qualitative experimental results demonstrate the superior performance of the proposed method compared with other state-of-the-art methods.
\end{itemize}

\section{Related Work}
\label{sec:related work}

\subsection{Personalized Image Generation}
Based on the impressive capabilities of large-scale text-to-image diffusion models \cite{rombach2022high, ramesh2022hierarchical, pernias2023wuerstchen}, personalized image generation has seen significant advancements in recent years. Existing methods can be divided into two categories according to whether fine-tuning is required for the input images. Early methods \cite{ruiz2023dreambooth, gal2022image, voynov2023p+, chen2024subject, tewel2023key, kumari2023multi, han2023svdiff, liu2023cones, hao2023vico, hu2021lora} require several example images of the specified subject to perform test-time optimization. Concretely, DreamBooth \cite{ruiz2023dreambooth} employs a rare token in the vocabulary to signify the given subject and fine-tunes the entire UNet weights to accommodate it. In contrast, Textual Inversion \cite{gal2022image} introduces a new token into the vocabulary to represent the subject. To reduce the optimization costs, Custom Diffusion \cite{kumari2023multi} fine-tunes only the key and value projection weights of the UNet. SVDiff \cite{han2023svdiff} performs parameter-efficient fine-tuning based on the singular-value decomposition of UNet weights. ViCo \cite{hao2023vico} introduces an image attention module to capture object-specific semantics. In addition, LoRA \cite{hu2021lora}, initially proposed for large language models, is adapted for personalized image generation by efficiently adjust the parameters in attention layers. However, fine-tuning-based methods typically require tens of minutes for a single subject, which is infeasible in practical applications.

More recently, some tuning-free methods \cite{wei2023elite, gal2023encoder, xiao2023fastcomposer, li2024blip, ma2023subject, shi2023instantbooth, arar2023domain, ruiz2023hyperdreambooth, ye2023ip, li2023photomaker} are proposed to enable instantaneous personalization in inference phase by training a vision encoder along with the diffusion model on large-scale datasets. ELITE \cite{wei2023elite}, trained on OpenImage testset \cite{kuznetsova2020open}, features a dual mapping network for the input image to align textual and image embeddings, consequently improving editability and detail. Focused on portrait generation, FastComposer \cite{xiao2023fastcomposer} employs subject embeddings inverted from the input reference image to augment the text conditioning and fine-tunes the entire UNet weights on the FFHQ-wild dataset \cite{karras2019style}. BLIP-Diffusion \cite{li2024blip} adopts a two-stage training approach for the diffusion model using the OpenImage dataset \cite{kuznetsova2020open}. IP-Adapter \cite{ye2023ip} introduces the decoupled the cross-attention strategy for image prompt adaptation, and trains the cross-attention layers on the LAION-2B \cite{schuhmann2022laion} and COYO-700M \cite{kakaobrain2022coyo-700m} datasets. PhotoMaker \cite{li2023photomaker} takes a similar approach to FastComposer \cite{xiao2023fastcomposer}, but stacks the embeddings of multiple reference images for enhancing ID preservation capabilities. In contrast, our method is generally applicable to various data domains and is able to generate high-fidelity images of both single- and multiple subjects.

\subsection{Multi-Subject Image Generation}
Custom Diffusion \cite{kumari2023multi}, as a pioneering multi-subject generation method, facilitates the multi-subjects generation by jointly training on the images of multiple subjects. In contrast, Cones \cite{liu2023cones} concatenates the concept neurons of multiple personalized models directly to enable multi-subject image generation. However, these methods encounter the attribute binding problem \cite{gu2024mix, xiao2023fastcomposer} when dealing with subjects of similar categories, resulting in the repeated synthesis of the same subject when synthesizing different subjects. To alleviate this problem, subsequent works propose some strategies to modulate the cross-attention maps. Mix-of-Show \cite{gu2024mix} utilizes both global and multiple regional prompts to describe an image, and merges the attention maps of different prompts for regionally controllable sampling. To alleviate the mutual interference between different subjects, Cones2 \cite{liu2023cones2} utilizes predefined layouts to rectifies the activations in cross-attenntion maps. Another line of methods \cite{han2023svdiff, xiao2023fastcomposer, ma2023subject} constrain the cross-attention maps during training for multi-subject image generation. FastComposer \cite{xiao2023fastcomposer} proposes cross-attention localization loss to ensure the attention of different subjects on appropriate regions. Following a similar approach as FastComposer \cite{xiao2023fastcomposer}, Subject-diffusion \cite{ma2023subject} employs an attention control mechanism to support multi-subject generation. Our MM-Diff similarly imposes constrains on cross-attention maps during the training phase, but simultaneously takes into account the interaction of text and image embeddings with the diffusion latents, which guarantees the generation of high-fidelity multi-subject images without any predefined input.

\section{Preliminaries}
\label{sec:preliminaries}

In this section, we provide an overview of the state-of-art text-to-image method, Stable Diffusion \cite{rombach2022high}, and parameter-efficient fine-tuning method, LoRA \cite{hu2021lora}, both of which serve as the basis of our method.

\subsection{Stable Diffusion} Stable Diffusion \cite{rombach2022high} conducts the diffusion process in the low-dimensional latent space to reduce computational costs. It comprises three components: a variational autoencoder (VAE) \cite{kingma2013auto} for compressing the input image into a latent representation, a text encoder for encoding the text prompt into embeddings, and a UNet \cite{ronneberger2015u} architecture for the denoising process. Specifically, given an input training image $x$, the VAE encoder $\varepsilon $ first transform it into a compressed latent representation $z$, which is then perturbed with a Gaussian noise $\epsilon$. The UNet, denoted as $\epsilon _{\theta }$, is employed to predict the noise $\epsilon$ conditioned on the text embeddings $c$. The training objective is defined as:

\begin{equation}
  \mathcal{L}_{SD}=\mathbb{E}_{\varepsilon \left ( x \right ) , c, \epsilon \sim \mathcal{N} \left ( 0,1 \right ) , t }\left [ \left \| \epsilon -\epsilon _{\theta \left ( z_{t}, t, c  \right ) }  \right \|_{2}^{2}   \right ],
  \label{eq:sd_loss}
\end{equation}
where $z_{t}$ represent the latents at timestep $t$. During inference, a random Gaussian noise is iteratively denoised by the UNet in the latent space, and the final image is obtained by decoding the latents with the VAE decoder.

Cross-attention mechanism is adopted in Stable Diffusion model to inject text information into the image generation process. Specifically, the latents $z$ and text embeddings $c$ are first projected using linear layers to obtain the query $Q=W^{q} z$, key $K=W^{k} c$ and value $V=W^{v} c$, and then the attention output is computed through a weighted sum over the value features, which can be formulated as:

\begin{equation}
  {\rm Attention}\left ( Q, K, V \right ) = {\rm Softmax} \left ( \frac{QK^{T} }{\sqrt{d} }  \right ) V,
  \label{eq:cross_attention}
\end{equation}
where $W^q$, $W^k$, $W^v$ are weight matrices of projection layers, and $d$ is the dimension of query and key features. The latents is then updated with the outputs of attention operation.

\subsection{Low-Rank Adaptation} Low-Rank Adaptation (LoRA) \cite{hu2021lora} was introduced to text-to-image related tasks for parameter-efficient fine-tuning. It hypothesizes that the weight updates during model adaptation have a low intrinsic rank and employs a low-rank decomposition of weight change to obtain the updated weight $W$, which is formulated as $W = W_0 + \bigtriangleup W = W_0 + BA$. Where $W_0 \in \mathbb{R} ^{d\times k} $ is the pretrained weight matrix which is frozen during adaption, $B \in \mathbb{R}^{d\times r}$ and $A \in \mathbb{R}^{r\times k}$ are the low-rank matrices containing trainable parameters, $r\ll \min \left ( d, k \right ) $ is the rank.

\section{Method}
\label{sec:method}

\begin{figure}[tb]
  \centering
  \includegraphics[width=12.0cm]{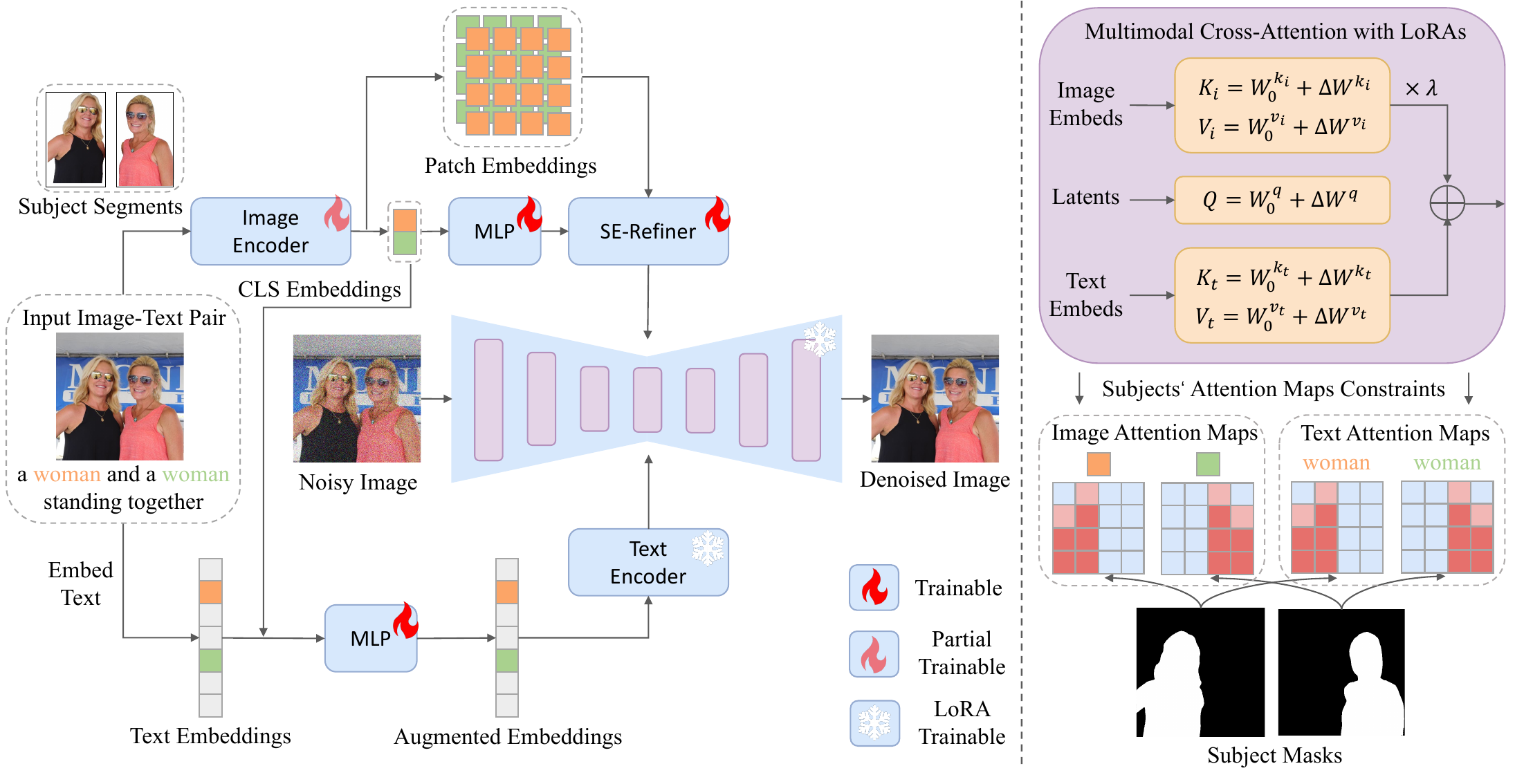}
  \caption{The overall pipeline of the proposed MM-Diff. On the left, the vision-augmented text embeddings and a small set of detail-rich subject embeddings are injected into the diffusion model through the well-designed multi-modal cross-attention. On the right, we illustrate the details of the innovative implementation of cross-attention with LoRAs, as well as the attention constraints that facilitate multi-subject generation.
  }
  \label{fig:pipeline}
\end{figure}

\subsection{Overview}
Given a reference image containing the subject to be customized, the proposed MM-Diff aims to to generate a new image depicting the specified subject in various scenes, styles and actions by the text prompt. Specifically, MM-Diff employs a CLIP vision encoder \cite{radford2021learning} to transform the visual subject into global embeddings (\ie, CLS embeddings) and patch-level embeddings. To enhance text and subject consistency simultaneously, the CLS embeddings, on the one hand, are used to augment text embeddings, and on the other hand, are used to derive a small number of detail-rich subject embeddings by the proposed Subject Embedding Refiner (SE-Refiner). These refined subject embeddings, along with the augmented text embeddings, are then injected into the diffusion model through the innovative application of LoRA layers \cite{hu2021lora}. Furthermore, we introduce cross-attention map constraints to tackle attribute binding problem in multi-subject generation. These constraints guide the model to associate entity tokens in both text and image condition with unique image regions during training, thereby allowing for flexible generation of multi-subject images without any predefined input (\eg, layout) during inference. The overall pipeline of the proposed method is illustrated in \cref{fig:pipeline}.

\subsection{Image Generation with Multi-Modal Condition Injection}
\label{sec:image generation}
Compared to text-to-image generation which relies solely on the text condition, personalized image generation also incorporates the image condition. For improving subject fidelity, previous methods typically involve direct fine-tuning on the base model or the injection of dense visual embeddings into it, both of which are inefficient and result in poor generalization. To address these limitations, we introduce a novel architecture consisting of four key components.

\subsubsection{Image Encoder.} Following recent works \cite{xiao2023fastcomposer, ye2023ip}, we employ a pretrained CLIP \cite{radford2021learning} image encoder to extract visual embeddings of the reference image. We utilize both the CLS embeddings and patch-level embeddings, where the former exhibits strong alignment with text embeddings, and the latter encapsulates rich subject details. To improve text consistency of the generated images, we use CLS embeddings of the subject to provide auxiliary information to augment the text condition. To obtain detail-rich visual condition, a MLP is utilized to project CLS embeddings into a small number of subject embeddings, which are then enriched with patch embeddings through the proposed SE-Refiner. In the training stage, we mask the subject's background with random noise to mitigate overfitting on irrelevant elements. To accommodate these artificially altered input images, part of the transformer layers are fine-tuned.

\subsubsection{Vision-Augmented Text Embeddings.} Effectively integrating image and text condition is crucial for enhancing the subject fidelity and text editability of the generated images. Similar to the approach in \cite{ma2023subject}, we augment the text condition with image condition before the text encoder, but retain original text embeddings to provide class priors. Specifically, for the given text prompt, we first get the word embeddings via lookup process. Next, we concatenate the word embeddings of entity (\eg, man) with the corresponding CLS embeddings and feed the augmented text embeddings into a MLP for fusion. The resulting embeddings are used as the input of the text encoder to get the final text embeddings. This process is illustrated in \cref{fig:pipeline}. We add LoRA layers to all attention layers in text encoder to adapt to the changes of inputs.

\subsubsection{Subject Embedding Refiner.} To improve the subject fidelity between the generated images and the reference image, a straightforward approach might involve injecting dense patch embeddings into the UNet model, similar to the techniques described in \cite{shi2023instantbooth, ma2023subject}. However, this approach can lead to a significant computational burden, as the length of the patch embeddings is typically quite extensive (\eg, 256). To tackle this problem, we opt for a more efficient approach by generating a small number of subject embeddings (\ie, 8) as the image condition. These subject subject embeddings, along with augmented text embeddings, are employed to control the content of the generated images. As describe before, the subject embeddings are derived from the subject CLS embeddings. We design the Subject Embedding Refiner (SE-Refiner) to enrich the details of subject embeddings using patch embeddings. This is achieved through a $N$-layer standard transformer decoder \cite{vaswani2017attention}. This design significantly improves the consistency of image subjects while maintaining a low computational overhead, as described in \cref{sec:ablation}.

\subsubsection{Multi-Modal Cross-Attention with LoRAs.} We incorporate LoRA layers \cite{hu2021lora} into all attention layers of the Stable Diffusion to integrate the vision-augmented text embeddings $c_t$, while preserving the generalizability of the original model. To inject subject embeddings $c_i$, another cross-attention with LoRAs is introduced. Specifically, we add a new set of LoRA layers for the key and value projections of each cross-attetnion layer in the UNet model. The outputs of text cross-attention and image cross-attention are combined to yield the updated latent representations. This process can be formulated as:

\begin{equation}
\begin{split}
  z_{out} &= {\rm Attention}(Q, K_t, V_t) + {\rm Attention}(Q, K_i, V_i), \\
where &\left\{\begin{matrix}
\begin{split} 
 &Q=\left ( W^q_0 + \bigtriangleup W^q \right ) z\\
 &K_t=\left ( W^{k_t}_0 + \bigtriangleup W^{k_t} \right ) c_t; V_t=\left ( W^{v_t}_0 + \bigtriangleup W^{v_t} \right ) c_t\\
 &K_i=\left ( W^{k_i}_0 + \bigtriangleup W^{k_i} \right ) c_i; V_i=\left ( W^{v_i}_0 + \bigtriangleup W^{v_i} \right ) c_i\\
\end{split}
\end{matrix}\right.
\end{split}
  \label{eq:decoupled cross_attention}
\end{equation}

As described in \cref{sec:preliminaries}, only $\bigtriangleup W$ is trainable.

\subsection{Multi-Subject Composition via Attention Map Constraints} Text-to-image models encounter attribute binding problem when generating images containing multiple entities, which means subjects of similar categories are mixed and those of different categories are incorrectly aligned with each other. Some works \cite{xiao2023fastcomposer, ma2023subject, gu2024mix} suggest that this problem stems from the unrestricted nature of the cross-attention mechanism, whereby a single token related to one subject can simultaneously attend to tokens associated with other subjects. Build on the insights from \cite{xiao2023fastcomposer, ma2023subject}, we introduce regularization terms to constrain the cross-attention maps, but additionally takes into account the incorporation of subject embeddings into the diffusion process. The core of our approach is to encourage entity tokens in both text and subject embeddings only to attend to the image region occupied by the corresponding subject during the training phrase, thus allowing for high-fidelity multi-subject generation.

\subsubsection{Text Cross-Attention Constraint.} The cross-attention mechanism in original Stable Diffusion scatters textual information across 2D latent features, ensuring that the generated images are consistent with the input text prompt. For each entity token within the text embeddings, we can obtain its attention map $A^{tca}$ of size $h \times w$, where h and w are the spatial size of the latent features. Intuitively, an entity token's attention should be focused exclusively on its corresponding image region, rather than on regions belonging to other entities, to prevent attribute binding issue. To achieve this, we introduce a regularization term that penalizes the L1 deviation between the entity's attention maps and the corresponding segmentation masks. Formally, the text cross-attention constraint can be formulated as:

\begin{equation}
  \mathcal{L}_{tcac}=\frac{1}{N}\sum_{i=1}^{N} \left | A^{tca}_i - M_i \right |,
  \label{eq:tcac_loss}
\end{equation}
where $A_i$ and $M_i$ represent $i$-th entity's attention map and the corresponding segmentation mask.

\subsubsection{Image Cross-Attention Constraint.} In addition to the text condition, the image condition (\ie, subject embeddings) is also introduced to improve subject consistency, as describe in \cref{sec:image generation}. For each subject embeddings of one entity, we could get its attention map $A^{ica}$ of size $h \times w$. Similarly, we introduce a regularization term that penalizes the L1 deviation between the subject embeddings' attention map and the corresponding entity's segmentation mask. Formally, the image cross-attention constraint can be formulated as:

\begin{equation}
  \mathcal{L}_{icac}=\frac{1}{MN}\sum_{i=1}^{N}\sum_{j=1}^{M} \left | A^{ica}_{ij} - M_i \right |,
  \label{eq:ica_loss}
\end{equation}
where $A_{ij}$ represents $i$-th entity's $j$-th subject embeddings' attention map.

The final training objective of the proposed method is:

\begin{equation}
  \mathcal{L} = L_{SD} + \lambda_{tcac} \mathcal{L}_{tcac} + \lambda _{icac} \mathcal{L}_{icac},
  \label{eq:total_loss}
\end{equation}
where $\lambda_{tcac}$ and $\lambda _{icac}$ are the hyperparameters that control the influence of the regularization constraints.

\section{Experiments}
\subsection{Experimental Settings}

\begin{figure}[tb]
  \centering
  \includegraphics[width=12.0cm]{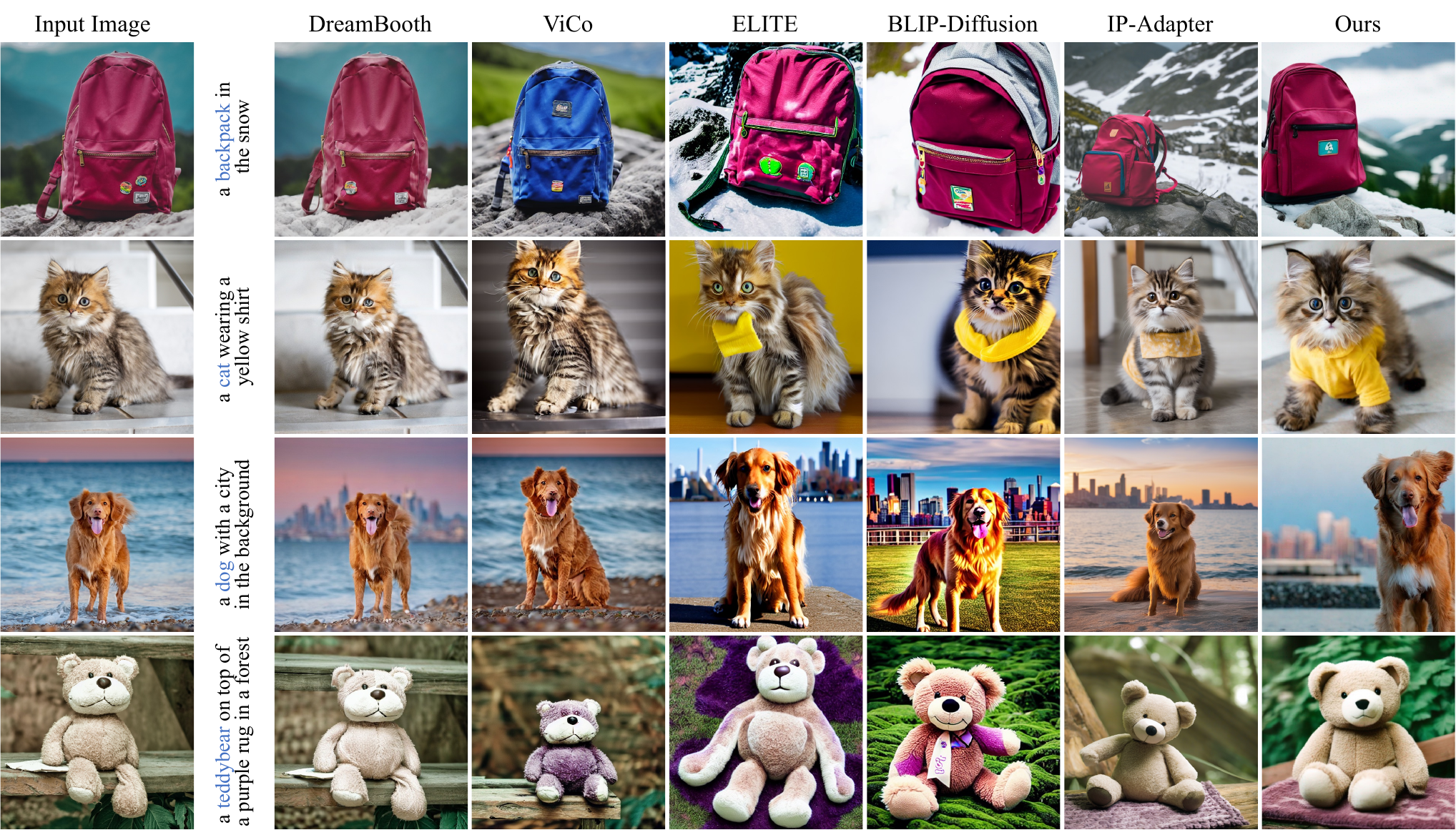}
  \caption{Visual comparisons on single subject generation.
  }
  \label{fig:general_subjects}
\end{figure}

\subsubsection{Datasets.} We construct a structured dataset based on ImageNet-1k \cite{russakovsky2015imagenet} dataset to train our model, comprising 280k images spanning 1k object classes. This dataset encompasses various modalities, including image-text pairs, entity boxes and masks. The data construction rules are similar to \cite{xiao2023fastcomposer, ma2023subject}. For portrait generation, we directly adopt the FFHQ-wild \cite{karras2019style} dataset processed by \cite{xiao2023fastcomposer} to train our model, which consists of 70k images with varying numbers of entities.

\subsubsection{Training Details.} We adopt Stable Diffusion-XL\footnote{https://huggingface.co/stabilityai/stable-diffusion-xl-base-1.0} as our base model, but the resolution of training data is resized to $512 \times 512$ due to resource constraints. For encoding reference images, we leverage OpenAI's clip-vit-large-patch14\footnote{https://huggingface.co/openai/clip-vit-large-patch14} vision model, which is aligned with one of the text encoders in Stable Diffusion-XL. During training, we freeze the original weights of base model, and only train the MLPs, LoRA layers in the text encoders and UNet, as well as the final four transformer blocks of the vision encoder. We train our model for 300k steps on two NVIDIA A100 GPUs, with a batch size of 8 per GPU. We apply a constant learning rate of 1e-4 for UNet LoRA layers and 1e-5 for the remaining trainable parameters. The constraint hyperparameters, $\lambda_{tcac}$ and $\lambda_{icac}$, are all set to 0.001. A maximum of two reference subjects are set during training, with 10$\%$ chance for each subject to be dropped. To enable classifier-free guidance sampling, the model is trained unconditionally with probability 10$\%$. During inference, We use 25-step DPM-solver \cite{lu2022dpm} sampler with a guidance scale of 5.

\subsubsection{Evaluation Metric.} We evaluate our method using subject fidelity and text fidelity metrics. For general subject generation, subject fidelity is measured by the CLIP image similarity (CLIP-I) and DINO \cite{caron2021emerging} similarity between the generated images and reference images. For portrait generation, subject fidelity is evaluated by face similarity, which we compute using FaceNet \cite{schroff2015facenet} on the cropped facial regions detected by RetinaFace \cite{deng2020retinaface} between the generated images and reference images. Text fidelity is evaluated by the CLIP text-image similarity (CLIP-T) between the generated images and text prompts for all types of datasets.

\subsection{Single-Subject Image Generation}

\begin{figure}[tb]
  \centering
  \includegraphics[width=12.0cm]{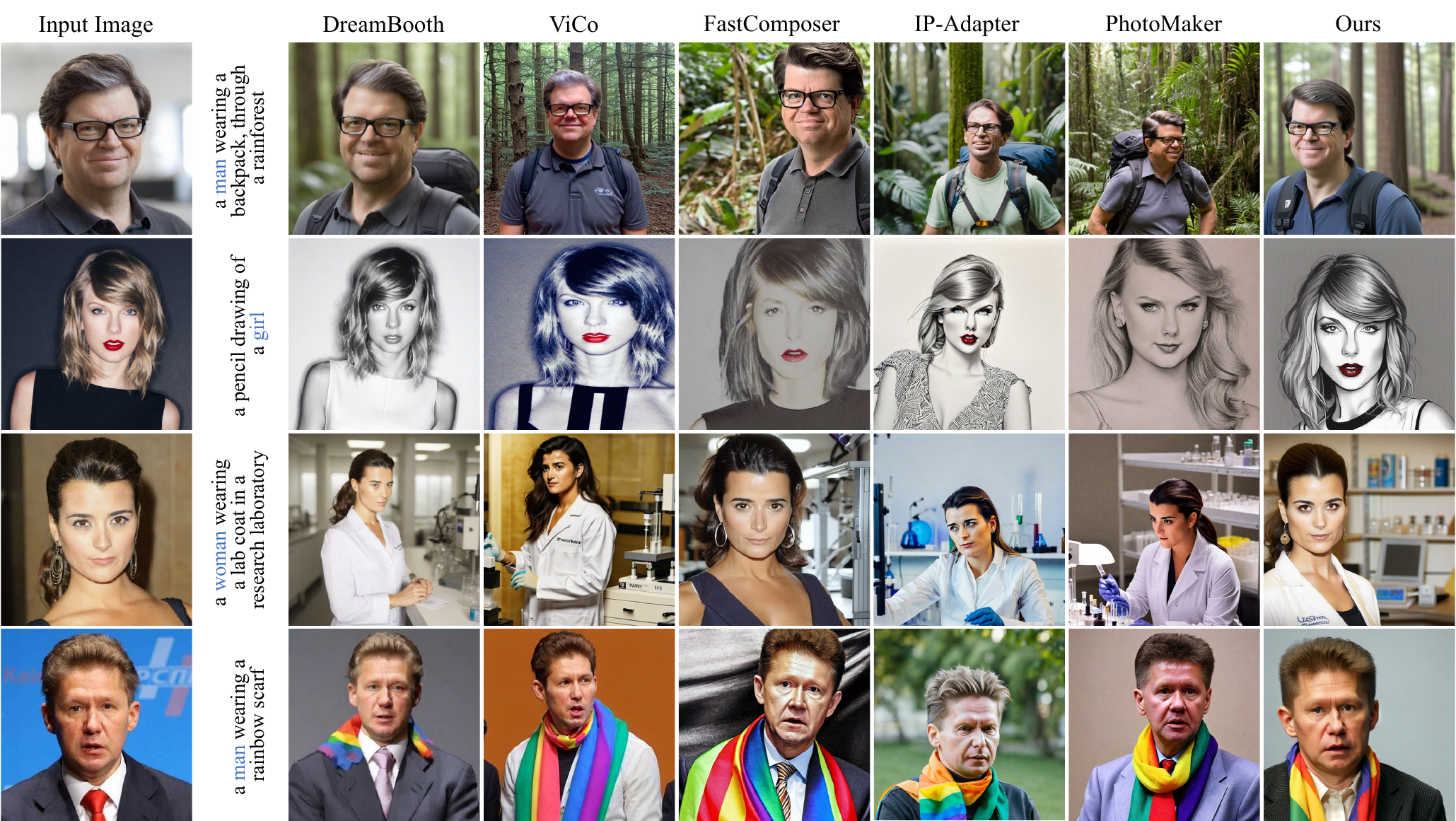}
  \caption{Visual comparisons on portrait generation.
  }
  \label{fig:single_subject_human}
\end{figure}

\begin{figure}[tb]
  \centering
  \includegraphics[height=7.5cm]{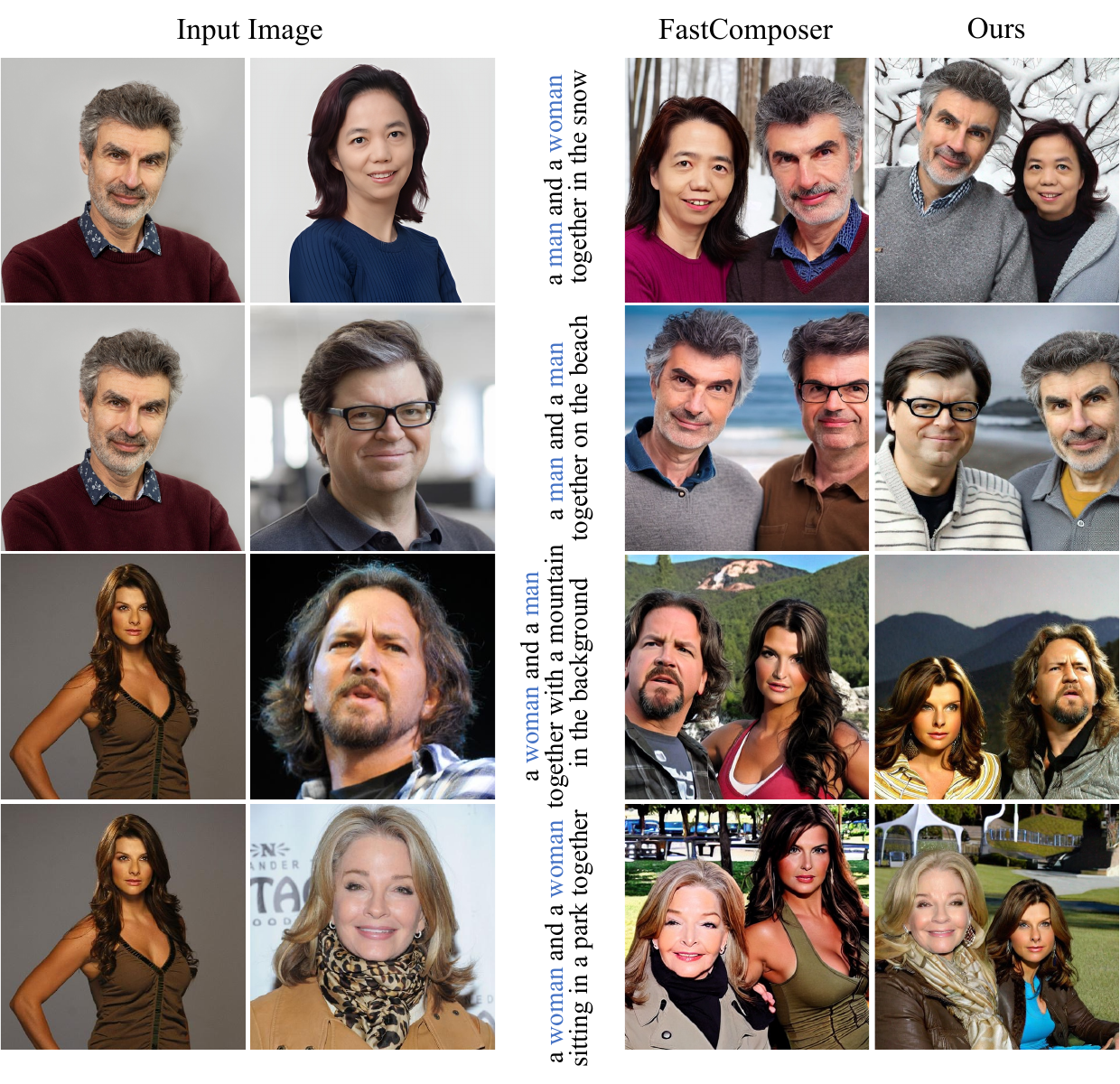}
  \caption{Visual comparisons on multi-subject generation.
  }
  \label{fig:multi_subjects}
\end{figure}

\subsubsection{Comparison Results for General Subject Generation.} We compare our MM-Diff, trained on the subset of ImageNet-1k \cite{russakovsky2015imagenet}, against both fine-tuning methods, including DreamBooth \cite{ruiz2023dreambooth} and ViCo \cite{hao2023vico}, and tuning-free methods, including ELITE \cite{wei2023elite}, BLIP-Diffusion \cite{li2024blip} and IP-Adapter \cite{ye2023ip}. For DreamBooth and IP-Adapter, we employ their Stable Diffusion-XL variants to ensure a fair comparison. We evaluate the performance on the subjects drawn from DreamBench \cite{ruiz2023dreambooth} benckmark. The entire test dataset consists of 15 subjects, each paired with 20 diverse text prompts. In \cref{tab:general subjects}, we generate 6 images per text prompt, yielding a total of 1800 images for comparison. We present the average CLIP-I, DINO, and CLIP-T scores over all reference and generated image pairs. The quantitative results demonstrate that our method achieves better subject fidelity and comparable text consistency relative to other leading tuning-free methods. \cref{fig:general_subjects} shows the qualitative results under various prompts using different methods. We can see that while fine-tuning methods such as Dreambooth achieve higher subject fidelity, they tend to generate repetitive and monotonous images. In contrast, our method not only retains the subject details but also excels in generating diverse images.

\subsubsection{Comparison Results for Portrait Generation.} We compare our MM-Diff, trained on FFHQ-wild \cite{karras2019style}, with both fine-tuning methods, including DreamBooth \cite{ruiz2023dreambooth} and ViCo \cite{hao2023vico}, and tuning-free portrait generation methods, including FastComposer \cite{xiao2023fastcomposer}, IP-Adapter-FaceID \cite{ye2023ip} and PhotoMaker \cite{li2023photomaker}. We utilize the Stable Diffusion-XL variants of DreamBooth, IP-Adapter-FaceID and PhotoMaker to ensure a fair comparison. Similar to the protocol in \cite{xiao2023fastcomposer}, we compose our evaluation benchmark from Celeb-A dataset \cite{liu2015deep}, which consists of 10 subjects, each paired with 20 unique text prompts. We generate 6 images per text prompt, amounting in a total of 1200 images for comparative analysis. As shown in \cref{tab:person subjects}, MM-Diff surpasses all comparison methods in face similarity. However, in terms of text consistency, our method does not perform as well as PhotoMaker, which we attribute to the limited size of our training dataset. \cref{fig:single_subject_human} presents visual comparison examples.

\begin{table}[tb]
  \caption{Quantitative comparisons on single subject generation.}
  \label{tab:general subjects}
  \centering
  \begin{tabular}{@{}lccccc@{}}
    \toprule
    Methods & Type & CLIP-I$\uparrow$ ($\%$) & DINO$\uparrow$ ($\%$) & CLIP-T$\uparrow$ ($\%$) & Speed$\downarrow$ (s)\\
    \midrule
    DreamBooth \cite{ruiz2023dreambooth} & Finetune & \textbf{93.1} & 79.2 & 25.7 & 507\\
    ViCo \cite{hao2023vico} & Finetune & 80.1 & 67.6 & 23.7 & 1226\\
    \cline{1-6}
    ELITE \cite{wei2023elite} & Zero shot & 83.5 & 71.2 & 22.8 & 3\\
    BLIP-Diffusion \cite{li2024blip} & Zero shot & 83.1 & 68.4 & 29.4 & \textbf{2}\\
    IP-Adapter \cite{ye2023ip} & Zero shot & 87.4 & 70.9 & \textbf{29.9} & 12\\
    MM-Diff (ours) & Zero shot & 91.5 & \textbf{80.8} & 27.8 & 4\\
  \bottomrule
  \end{tabular}
\end{table}

\begin{table}[tb]
  \caption{Quantitative comparisons on portrait generation.}
  \label{tab:person subjects}
  \centering
  \begin{tabular}{@{}lccccc@{}}
    \toprule
    Methods & Type & Face Sim.$\uparrow$ ($\%$) & CLIP-T$\uparrow$ ($\%$) & Speed$\downarrow$ (s)\\
    \midrule
    DreamBooth \cite{ruiz2023dreambooth} & Finetune & 62.6 & 25.8 & 436\\
    ViCo \cite{hao2023vico} & Finetune & 39.1 & 21.4 & 1241\\
    \cline{1-5}
    FastComposer \cite{xiao2023fastcomposer} & Zero shot & 57.5 & 26.6 & 4\\
    IP-Adapter-FaceID \cite{ye2023ip} & Zero shot & 61.0 & 30.6 & 14\\
    PhotoMaker \cite{li2023photomaker} & Zero shot & 49.3 & \textbf{32.7} & 15\\
    MM-Diff (ours) & Zero shot & \textbf{75.6} & 27.0 & \textbf{4}\\
  \bottomrule
  \end{tabular}
\end{table}

\subsection{Multi-Subject Image Generation} We conduct experiments on multi-subject image generation. In the absence of open-source methods for general subject, we compare our method with FastComposer \cite{xiao2023fastcomposer} for portrait generation. To evaluate multi-subject generation performance, we utilize all possible combinations (45 pairs in total) of the previously described 10 person subjects, with each paired with 10 distinct text prompts. \cref{tab:multi-subject} presents the quantitative comparison results, which exhibit the superior performance of the proposed method over FastComposer, especially in terms of face similarity. The qualitative comparisons, as illustrated in \cref{fig:multi_subjects}, further demonstrate the effectiveness of our method in generating high-fidelity multi-subject images.

\subsection{Ablation Study}
\label{sec:ablation}
We conduct ablation studies to evaluate the effects of key components in our method, including the vision-augmented text embeddings, the SE-Refiner, and the cross-attention map constraints. \cref{tab:ablation} presents the ablation results for single subject generation, which indicate that the removal of any of these components leads to a decline in performance compared to the full setting baseline.

\subsubsection{Effect of Vision-Augmented Text Embeddings.} \cref{tab:ablation} presents the ablation studies examining the effect of vision-augmented text embeddings. For comparison, we introduce a variant where the image-text fusion process is removed and the original text embeddings are directly injected into the UNet. Experimental results demonstrate that the use of vision-augmented text embeddings leads to a superior performance on both metrics.

\subsubsection{Effect of SE-Refiner.} \cref{tab:ablation} presents the ablation studies on the SE-Refiner. We introduce a variant for comparison where the SE-Refiner is removed and the original subject embeddings are instead directly injected into the UNet. Experimental results suggest that the absence of SE-Refiner leads to a marked decline in terms of subject fidelity, primarily attributable to the loss of details.

\begin{table}[tb]
  \caption{Quantitative comparisons on multi-subject generation.}
  \label{tab:multi-subject}
  \centering
  \begin{tabular}{@{}lccccc@{}}
    \toprule
    Methods & Type & Face Sim.$\uparrow$ ($\%$) & CLIP-T$\uparrow$ ($\%$) & Speed$\downarrow$ (s)\\
    \midrule
    FastComposer \cite{xiao2023fastcomposer} & Zero shot & 54.0 & 26.6 & 4\\
    MM-Diff (ours) & Zero shot & \textbf{65.5} & \textbf{27.1} & \textbf{4}\\
  \bottomrule
  \end{tabular}
\end{table}

\begin{table}[tb]
  \caption{Ablation studies of key components of our method.}
  \label{tab:ablation}
  \centering
  \begin{tabular}{@{}lccc@{}}
    \toprule
    Methods & CLIP-I$\uparrow$ ($\%$) & DINO$\uparrow$ ($\%$) & CLIP-T$\uparrow$ ($\%$)\\
    \midrule
    Baseline & \textbf{91.5} & \textbf{80.8} & \textbf{27.8}\\
    \cline{1-4}
    w/o vision augmentation & 84.7$\downarrow$ & 71.1$\downarrow$ & 27.2$\downarrow$\\
    w/o SE-Refiner & 76.3$\downarrow$ & 63.1$\downarrow$ & 27.5$\downarrow$\\
    w/o cross-attention map constraints & 87.3$\downarrow$ & 75.5$\downarrow$ & 27.4$\downarrow$\\
  \bottomrule
  \end{tabular}
\end{table}

\subsubsection{Effect of Cross-Attention Map Constraints.} \cref{tab:ablation} presents the ablation studies concerning the cross-attention map constraints. For comparison, we present a variant in which the cross-attention map constraints are removed while all other settings remain unchanged. Experimental results demonstrate that the proposed constraints not only enable multi-subject generation but also bring an obvious improvement on subject fidelity. This improvement can be attributed to the mechanism's ability to prevent the model from confusing different subjects, thus facilitating its convergence to a better solution.

\section{Conclusion}

We propose MM-Diff, a unified and tuning-free framework for image personalization. By efficiently integrating detail-rich subject embeddings along with vision-augmented text embeddings into the diffusion model, it is capable of generating high-fidelity images of both single and multiple subjects within seconds. Furthermore, MM-Diff tackles the attribute binding challenge in multi-subject generation through the well-designed cross-attention map constraints imposed during the training phase. This allows for flexible multi-subject image sampling during inference without any predefined inputs (\eg, layout). Experimental results on multiple testsets demonstrate that our method achieves superior performance than other leading methods.

\clearpage

\bibliographystyle{splncs04}
\bibliography{egbib}

\clearpage
\appendix

\begin{table}[tb]
  \caption{Text prompts for general subject generation. <class> represents class word corresponding to the subject, such as backpack, dog, cat, etc.}
  \label{tab_ablation:general subjects}
  \centering
  \resizebox{\linewidth}{!}{
  \begin{tabular}{@{}ll@{}}
    \toprule
    Text prompts for non-live objects & Text prompts for animals \\
    \midrule
     a <class> in the jungle & a <class> in the jungle \\
     a <class> in the snow & a <class> in the snow \\
     a <class> on the beach & a <class> on the beach \\
     a <class> on a cobblestone street & a <class> on a cobblestone street \\
     a <class> on top of pink fabric & a <class> on top of pink fabric \\
     a <class> on top of a wooden floor & a <class> on top of a wooden floor \\
     a <class> with a city in the background & a <class> with a city in the background \\
     a <class> with a mountain in the background & a <class> with a mountain in the background \\
     a <class> on top of a purple rug in a forest & a <class> with a blue house in the background \\
     a <class> with a wheat field in the background & a <class> on top of a purple rug in a forest \\
     a <class> with a tree and autumn leaves in the background & a <class> wearing a Santa hat \\
     a <class> with the Eiffel Tower in the background & a <class> in a police outfit \\
     a <class> floating on top of water & a <class> wearing a yellow shirt \\
     a <class> on top of green grass with sunflowers around it & a <class> in a purple wizard outfit \\
     a <class> on top of a dirt road & a shiny <class> \\
     a <class> on top of a white rug & a wet <class> \\
     a red <class> & a cube shaped <class> \\
     a purple <class> & a <class> with Japanese modern city street in the background \\
     a shiny <class> & a <class> among the skyscrapers in New York city \\
     a cube shaped <class> & a <class> with a beautiful sunset \\
  \bottomrule
  \end{tabular}
  }
\end{table}

\begin{table}[tb]
  \caption{Text prompts for protrait generation. <class> represents class word corresponding to the subject, such as man, woman, girl, etc.}
  \label{tab_ablation:human subjects}
  \centering
  \resizebox{\linewidth}{!}{
  \begin{tabular}{@{}ll@{}}
    \toprule
    single person & multiple person \\
    \midrule
     a <class> wearing a santa hat & a <class> and a <class> together in the jungle \\
     a <class> wearing a rainbow scarf & a <class> and a <class> together in the snow \\
     a <class> wearing a black top hat and a monocle & a <class> and a <class> together on the beach \\
     a <class> wearing a yellow shirt & a <class> and a <class> together with a city in the background \\
     a <class> in the jungle & a <class> and a <class> together with a mountain in the background \\
     a <class> in the snow & a <class> and a <class> cooking a meal together \\
     a <class> on a cobblestone street & a <class> and a <class> sitting in a park together \\
     a <class> with a mountain in the background & a <class> and a <class> working out at the gym together \\
     a <class> with a city in the background & a <class> and a <class> baking cookies together \\
     a <class> holding a glass of wine & a <class> and a <class> gardening in the backyard together \\
     a <class> reading a book & - \\
     a <class> in a chef's outfit, cooking in a kitchen & - \\
     a <class> dressed in a a firefighter's outfit, a raging forest fire in the background & - \\
     a <class> wearing a casual t-shirt and jeans sits on a park bench, reading a book under a shady tree & - \\
     a <class> wearing a lab coat works diligently in a research laboratory, conducting scientific experiments to unlock new discoveries & - \\
     a <class> in an astronaut suit, floating in a spaceship & - \\
     a <class> dressed in bohemian festival attire, strums a guitar by a campfire under a starry sky & - \\
     a <class> adorned in traditional cultural clothing, gracefully practices a martial art in a serene garden & - \\
     a <class> wrapped in a cozy sweater, sits on a vintage armchair by a crackling fireplace, engrossed in a book & - \\
     a <class> wearing a backpack and adventure attire, confidently hikes through a lush rainforest & - \\
  \bottomrule
  \end{tabular}
  }
\end{table}

\section{Additional Implementation Details}
\subsection{Prompts for Evaluation} The text prompts used for evaluation are presented in \cref{tab_ablation:general subjects} and \cref{tab_ablation:human subjects}. These prompts are sourced from a combination of existing works \cite{ruiz2023dreambooth, hao2023vico, xiao2023fastcomposer}, alongside additional prompts generated by ChatGPT.

\subsection{Architecture of SE-Refiner} 
\begin{figure}[tb]
  \centering
  \includegraphics[height=6cm]{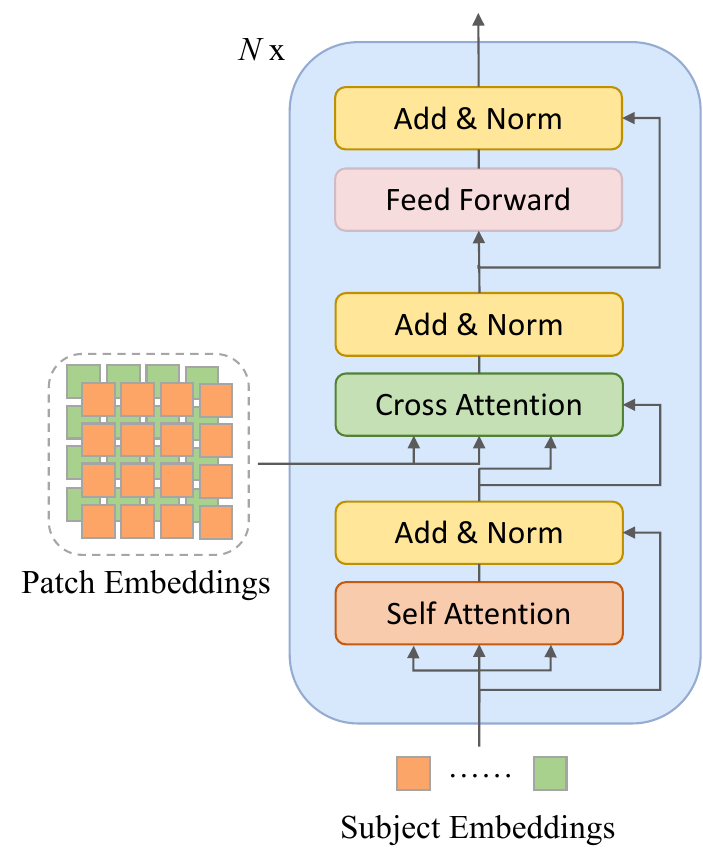}
  \caption{Architecture of SE-Refiner. It is implemented as a $N$-layer transformer decoder.
  }
  \label{fig_ablation:SE-Refiner}
\end{figure}

We show the detailed architecture of the SE-Refiner in \cref{fig_ablation:SE-Refiner}. It is designed as a decoder with four layers, following the standard transformer \cite{vaswani2017attention} structure. We first generate a sequence of subject embeddings from the CLIP CLS token of the reference image, which serve as the inputs to the SE-Refiner. Then, these embeddings are enriched with patch-level embeddings from the same CLIP vision encoder.

\section{Additional Ablation Studies}

\subsection{Effect of $\lambda$}
\begin{figure}[tb]
  \centering
  \includegraphics[width=12.0cm]{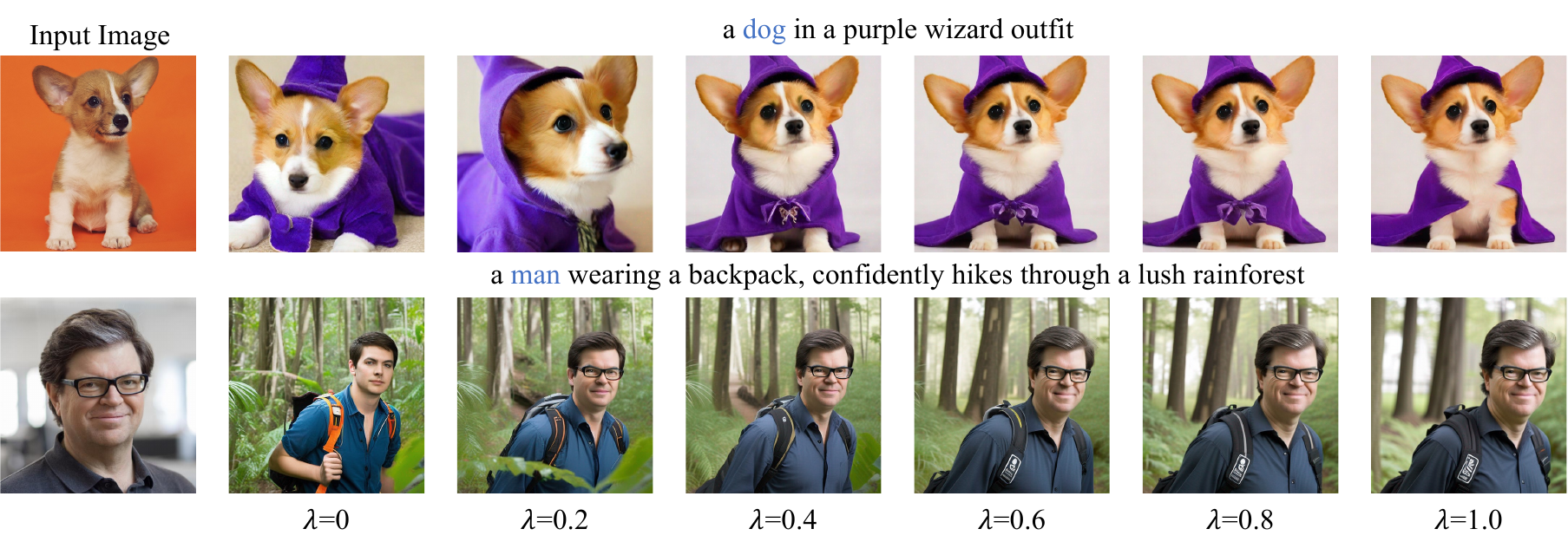}
  \caption{Effect of the parameter $\lambda$.
  }
  \label{fig_ablation:lambda}
\end{figure}

We conduct experiments to assess the effect of the parameter $\lambda$, which serves as a regulator for blending the text and image condition within the multi-modal cross-attention. To intuitively understand how $\lambda$ affects the generated results, we incrementally adjust its value within the range of 0 to 1. As shown in \cref{fig_ablation:lambda}, we can see that there is a positive correlation between the value of $\lambda$ and the subject fidelity. Nonetheless, it is also noted that excessively high values of $\lambda$ have the potential to impede the text consistency. Therefore, we set $\lambda = 0.8$ for trade-off, which delivers satisfactory results across a variety of scenarios.

\subsection{Visualization of Cross-Attention Maps} 
\begin{figure}
    \centering
    \begin{subfigure}{\linewidth} % 宽度设置为整个文本宽度
        \centering
        \includegraphics[height=3.5cm]{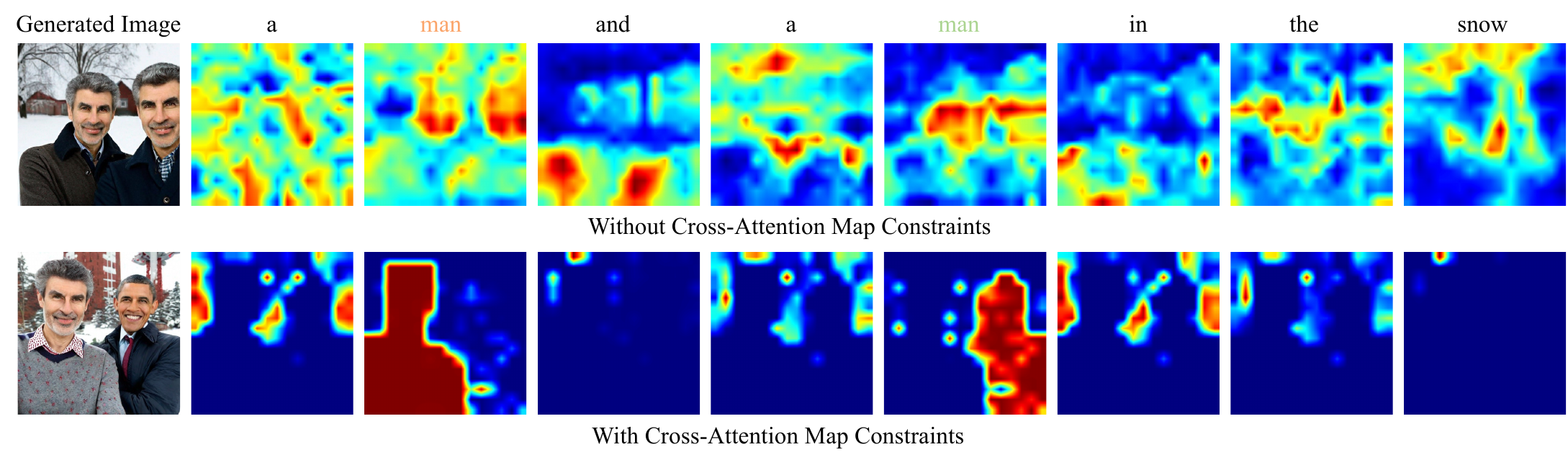}
        \caption{Visualization of text cross-attention maps.}
        \label{fig_ablation:text-attn-map}
    \end{subfigure}

    \begin{subfigure}{\linewidth}
        \centering
        \includegraphics[height=3.5cm]{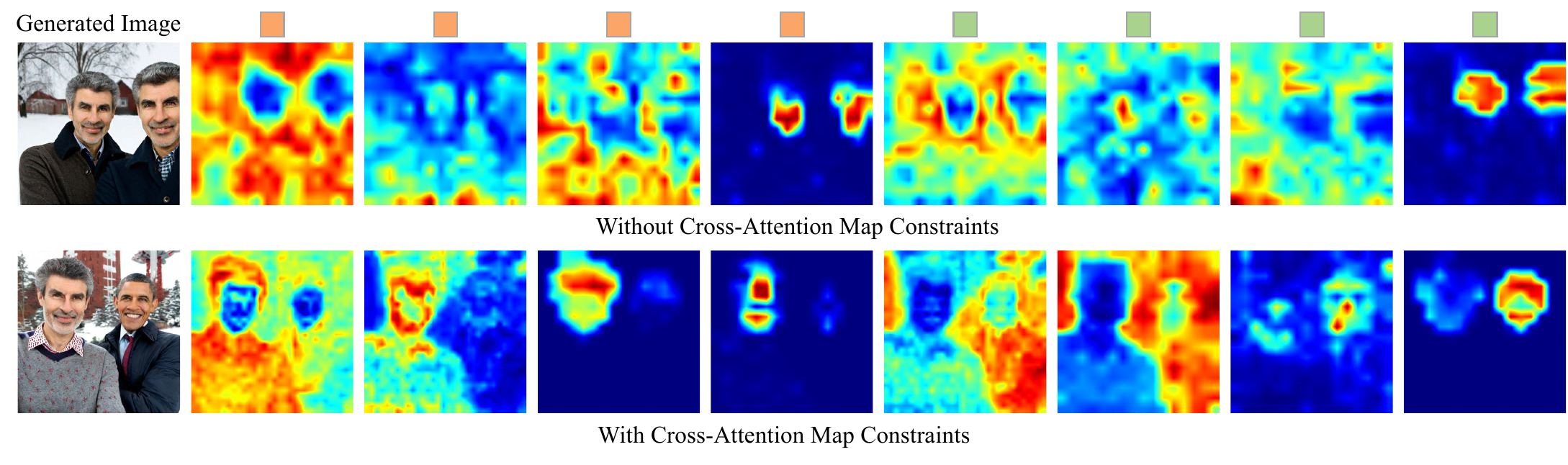}
        \caption{Visualization of image cross-attention maps.}
        \label{fig_ablatio :image-attn-map}
    \end{subfigure}
    \caption{Original model tends to mix the attributes of different subjects. By applying the proposed constraints, the model encourage entity tokens to focus exclusively on the corresponding image region, thus enabling multi-subject image generation.}
    \label{fig_ablation:attn_maps}
\end{figure}

In \cref{fig_ablation:attn_maps}, we demonstrate the effectiveness of the proposed cross-attention map constraints by showcasing the cross-attention maps in the UNet model. As we can see, the unconstrained model often allows multiple entity tokens to affect the same image region at once, leading to a mixture of attributes. With the proposed cross-attention constraints, the model is able to distinguish between the characteristics pertaining to different subjects.

\begin{figure}
    \centering
    \begin{subfigure}{\linewidth} % 宽度设置为整个文本宽度
        \includegraphics[width=\textwidth]{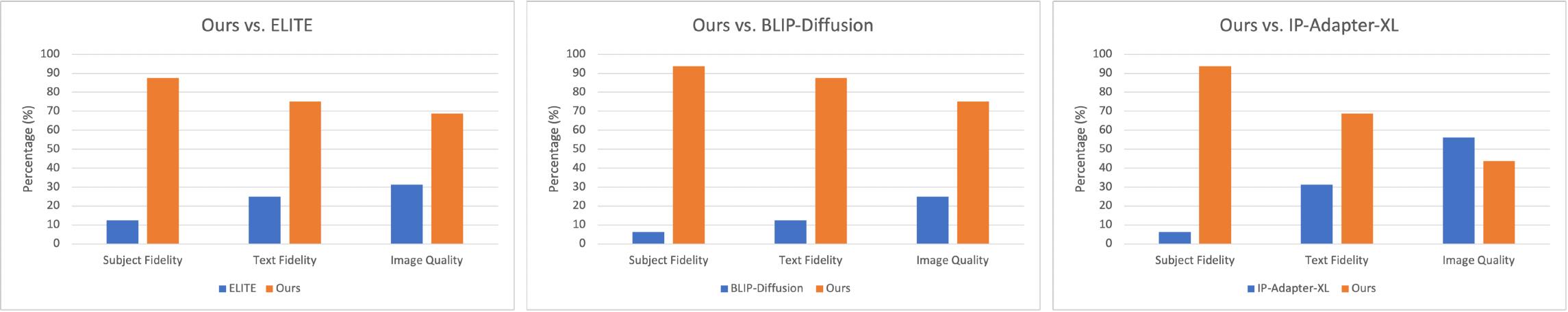}
        \caption{Percentage of user preferences on single subject generation.}
        \label{fig:user_study_single}
    \end{subfigure}
    \\

    \begin{subfigure}{\linewidth}
        \includegraphics[width=\textwidth]{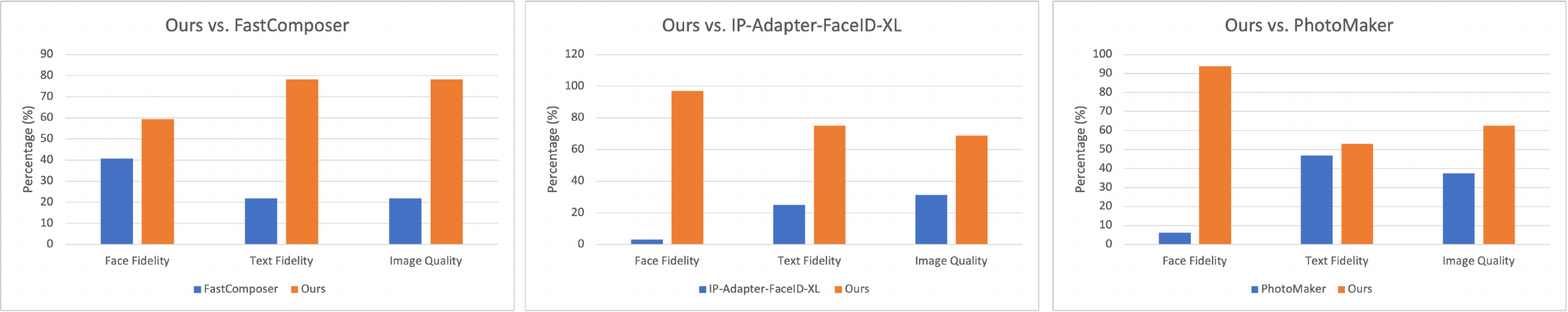}
        \caption{Percentage of user preferences on portrait generation.}
        \label{fig:user_study_portrait}
    \end{subfigure}
    \\

    \begin{subfigure}{\linewidth}
        \centering
        \includegraphics[width=4.0cm]{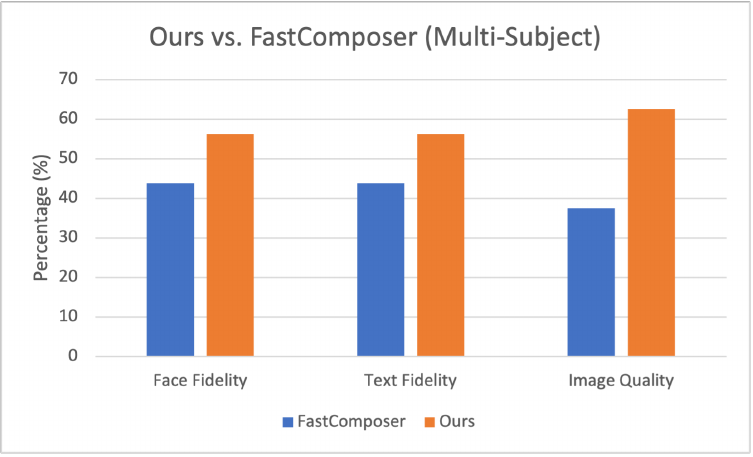}
        \caption{Percentage of user preferences on multi-subject generation.}
        \label{fig:user_study_multi}
    \end{subfigure}
    
    \caption{User studies on three criteria. Users tend to favor our method when compared to other methods.}
    \label{fig:user_study}
\end{figure}

\section{User Study}
We conduct user studies to comprehensively assess the quality of generated images. Since user study requires extensive human effort, we select top-tier tuning-free models for comparisons, including ELITE \cite{wei2023elite}, BLIP-Diffusion \cite{li2024blip}, IP-Adapter-XL \cite{ye2023ip}, FastCompoer \cite{xiao2023fastcomposer}, IP-Adapter-FaceID-XL \cite{ye2023ip}, and PhotoMaker \cite{li2023photomaker}. The evaluation is based on three criteria: 1) Subject/Face Fidelity - participants are shown a reference image alongside generated images from two different methods and are asked to identify the image that is more consistent with the reference. 2) Text Fidelity, participants are presented with a text prompt and generated images from two methods and are asked to select the image that aligns better with the text. 3) Image Quality, participants are shown generated images from two methods and are asked to select the one that appears aesthetically superior. According to the results shown in \cref{fig:user_study}, users tend to favor our method when compared to other methods.

\section{Limitations}
The training dataset of our model is relatively limited in size when benchmarked against other top-tier methods in the field. Also, the dataset used for general subject generation only contains one subject per image, which limits the model's multi-subject generation capabilities. To elevate the performance and broaden the generative scope of MM-Diff, integrating a more heterogeneous and expansive dataset is imperative.

\section{More Visualization Results of Portrait Generation}
\begin{figure}[tb]
  \centering
  \includegraphics[width=12.0cm]{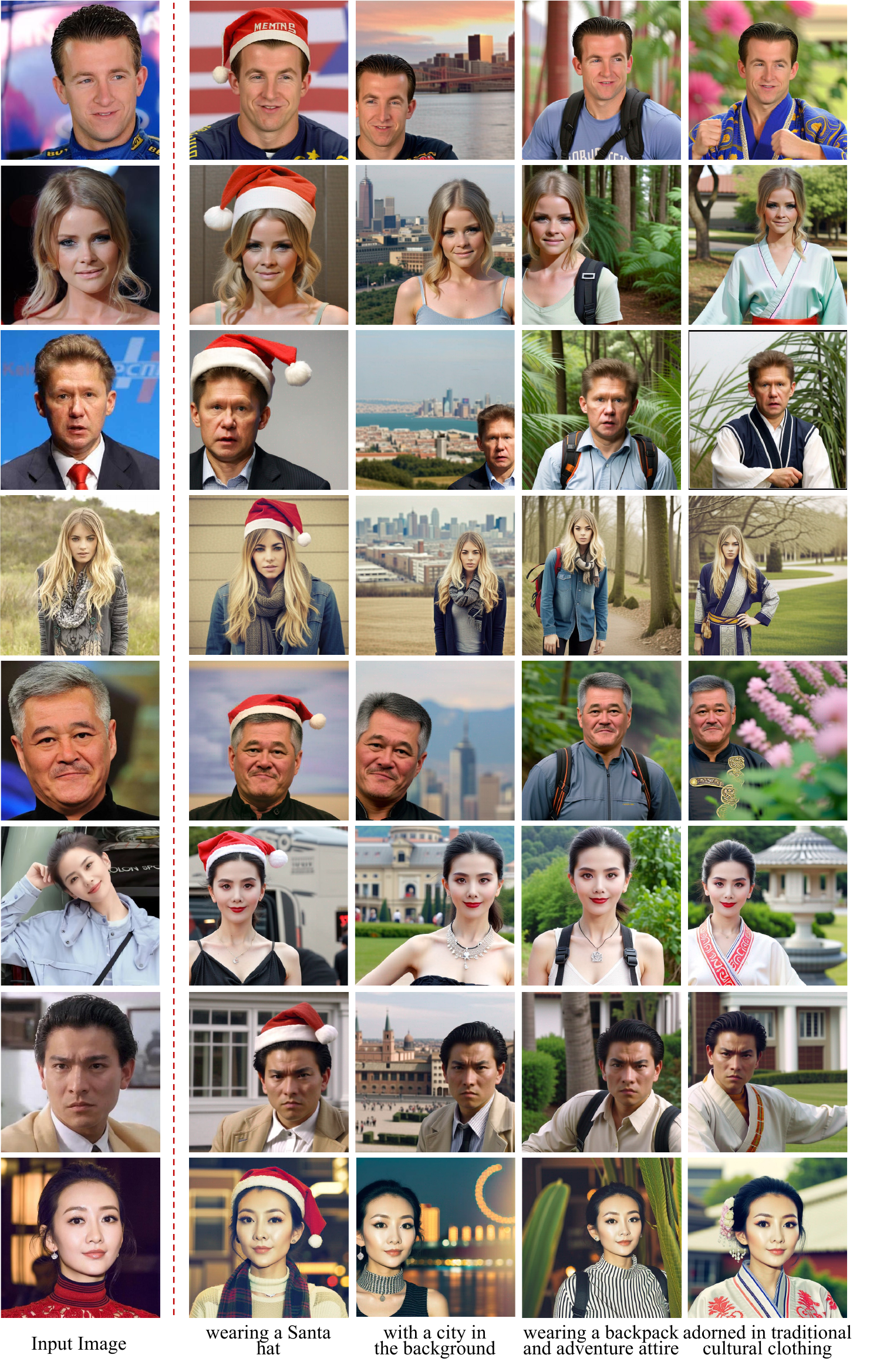}
  \caption{Additional visualization results of portrait generation.
  }
  \label{fig_more:human}
\end{figure}

\begin{figure}[tb]
  \centering
  \includegraphics[width=12.0cm]{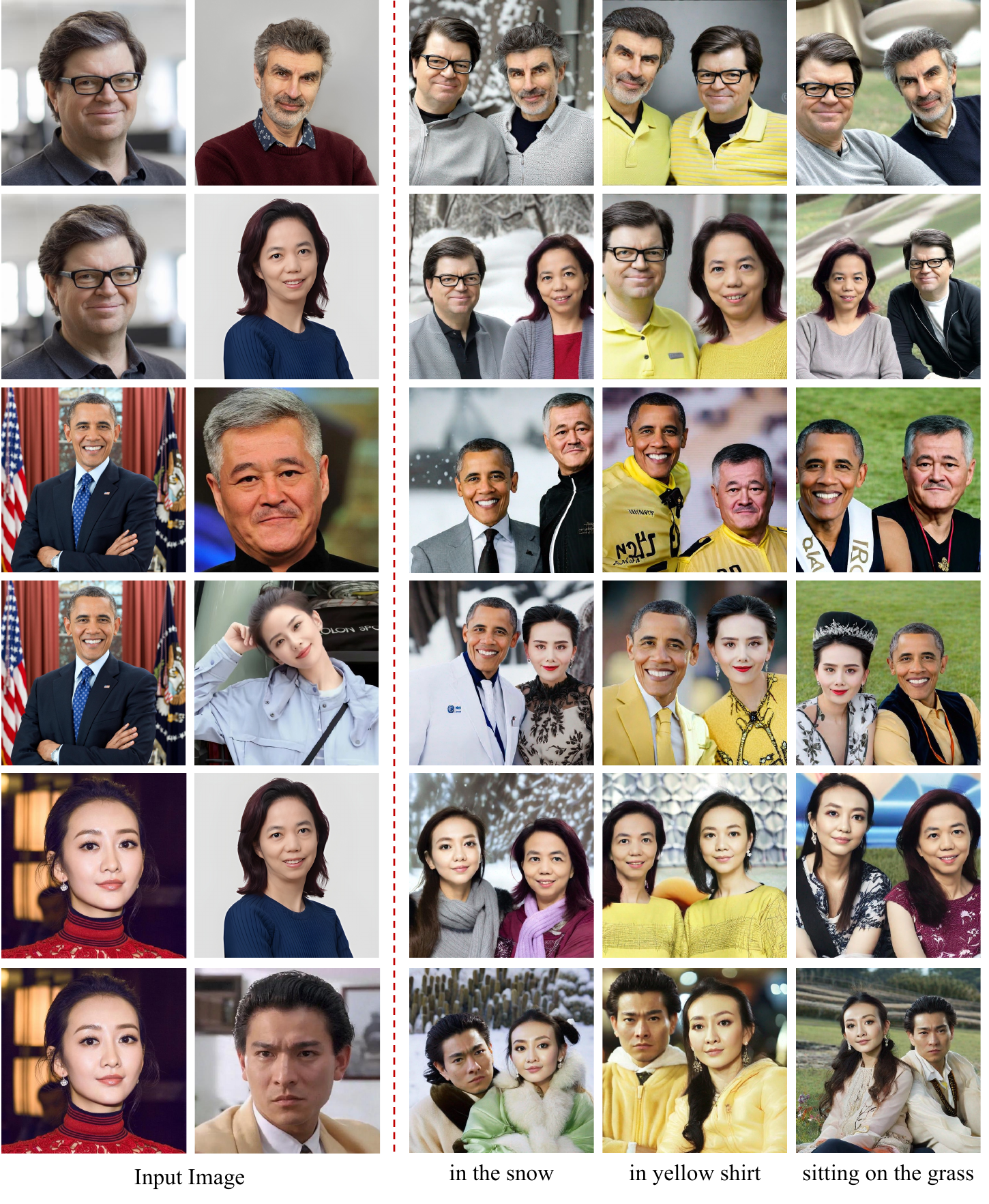}
  \caption{Additional visualization results of multi-subject generation.
  }
  \label{fig_more:multi_human}

\end{figure}

Recently, some portrait generation methods (\eg, PhotoMaker\cite{li2023photomaker}) have emerged, which have aroused great interest from the community. To further illustrate the generality and effectiveness of our method, we present more examples of portrait generation in \cref{fig_more:human} and \cref{fig_more:multi_human}.

\clearpage

\end{document}